\documentclass[11pt,a4paper]{article}
\usepackage[utf8]{inputenc}
\usepackage[hyperref]{emnlp-ijcnlp-2019}
\usepackage{times}
\usepackage{latexsym}
\usepackage{booktabs}
\usepackage{graphicx}
\usepackage{amsfonts}
\usepackage{amssymb}
\usepackage{url}
\makeatletter
\def\blfootnote{\xdef\@thefnmark{}\@footnotetext}
\makeatother
\aclfinalcopy 

\newcommand{\ours}{\textsc{football}}

\definecolor{mygreen}{rgb}{0,0.3,0.1}
\definecolor{myred}{rgb}{0.8,0.1,0.0}

\title{Investigating Sports Commentator Bias within a Large Corpus of American Football Broadcasts}

\author{\textbf{Jack Merullo}$^{\bigstar\spadesuit}$ \hspace{1cm}
        \textbf{Luke Yeh}$^{\bigstar\spadesuit}$ \hspace{1cm}
        \textbf{Abram Handler}$^{\spadesuit}$ \\
        \textbf{Alvin {Grissom II}}$^{\clubsuit}$ \hspace{1cm}
        \textbf{Brendan O'Connor}$^{\spadesuit}$ \hspace{1cm}
        \textbf{Mohit Iyyer}$^{\spadesuit}$
        \AND
        \normalfont University of Massachusetts Amherst$^{\spadesuit}$\hspace{7mm}
        \normalfont Ursinus College$^{\clubsuit}$\\
        \texttt{\{jamerullo,lyeh,ahandler,miyyer,brenocon\}@umass.edu}\\
        \texttt{agrissom@ursinus.edu}}

\date{}
\begin{document}
\maketitle
 \begin{abstract}
Sports broadcasters inject drama into play-by-play commentary by building team and player narratives through subjective analyses and anecdotes. Prior studies based on small datasets and manual coding show that such theatrics evince commentator bias in sports broadcasts. To examine this phenomenon, we assemble \ours, which contains 1,455 broadcast transcripts from American football games across six decades that are automatically annotated with 250K player mentions and linked with racial metadata. We identify major confounding factors for researchers examining racial bias in \ours, and perform a computational analysis that supports conclusions from prior social science studies.

\end{abstract}

\section{Introduction}
\label{sec:introduction}

Sports broadcasts are major events in contemporary popular culture: televised American football (henceforth ``football``) games regularly draw tens of millions of viewers
\citep{Palotta2019CNN}.
\blfootnote{$^{\bigstar}$\textnormal{Authors contributed equally.}}
Such broadcasts feature
live sports commentators
who 
weave the game's mechanical details
into a broader, more subjective narrative. 
Previous work suggests that this form of storytelling exhibits racial bias: nonwhite players are less frequently praised for good plays ~\cite{rainville1977extent}, while white players are more often credited with ``intelligence''~\cite{bruce2004marking, billings2004depicting}. However, such prior scholarship forms conclusions from small datasets\footnote{ \citet{rainville1977extent}, for example, study only 16 games.} and subjective manual coding of race-specific language.

We revisit this prior work using large-scale computational analysis.
From YouTube, we collect broadcast football transcripts and identify mentions of players, which we link
to metadata about each player's race and position. Our resulting \ours\ dataset contains over 1,400 games
spanning six decades, automatically annotated with 
$\sim$250K player mentions (Table~\ref{tab:dataset_examples}). Analysis of \ours\ identifies two confounding factors for research on racial bias: (1) the racial composition of many positions is very skewed (e.g., only $\sim$5\% of running backs are white), and (2) many mentions of players describe only their actions on the field (not player attributes). We experiment with an additive log-linear model for teasing apart these confounds. We also confirm prior social science studies on racial bias in naming patterns and sentiment. Finally, we publicly release \ours,\footnote{ \href{http://github.com/jmerullo/football}{\texttt{http://github.com/jmerullo/football}}} the first large-scale sports commentary corpus annotated with player race, to spur further research into characterizing racial bias in mass media.

\begin{table}[]
\small
\begin{tabular}{p{1.2cm}p{1.1cm}p{4cm}}
\toprule
\bf Player & \bf Race & \bf Mention text\\
\midrule
Baker\hspace{0.5cm}Mayfield & white & ``Mayfield the ultimate competitor he's tough he's scrappy''\\ \\
Jesse James & white & ``this is a guy \dots does nothing but work brings his lunch pail''\\
\midrule
Manny Lawson & nonwhite & ``good specs for that defensive end freakish athletic ability''\\ \\
B.J. Daniels & nonwhite & ``that otherworldly athleticism he has saw it with Michael Vick''\\
\bottomrule
\end{tabular}
\caption{Example mentions from \ours\ that highlight racial bias in commentator sentiment patterns.}
\label{tab:dataset_examples}
\vspace{-0.1in}
\end{table}

\section{Collecting the \ours\ dataset}
\label{sec:dataset}
We collect transcripts of 1,455 full game broadcasts from the
U.S.\ NFL and 
National Collegiate Athletic Association (NCAA) recorded between 1960 and 2019. Next, we identify and link mentions of players within these transcripts to information about their race (white or nonwhite) and position (e.g., quarterback). In total, \ours\ contains 267,778 mentions of 4,668 unique players, 65.7\% of whom are nonwhite.\footnote{See Appendix for more detailed statistics.} 
We now describe each stage of our data collection process.

\subsection{Processing broadcast transcripts}
We collect broadcast transcripts by downloading YouTube videos posted by nonprofessional, individual users identified by querying YouTube for football archival channels.\footnote{We specifically query for \texttt{full NFL$|$NCAA$|$college football games 1960s$|$1970s$|$1980s$|$1990s$|$2000}, and the full list of channels is listed in in the Appendix.}
YouTube automatically captions many videos, allowing us to scrape caption transcripts from 601 NFL games and 854 NCAA games. We next identify the teams playing and game's year by searching for exact string matches in the video title and manually labeling any videos with underspecified titles. 

After downloading videos, we tokenize transcripts using spaCy.\footnote{https://spacy.io/ (2.1.3), \citet{spacy2}} As part-of-speech tags predicted by spaCy are unreliable on our transcript text, we tag \ours\ using the ARK TweetNLP POS tagger~\cite{Owoputi2013ImprovedPT}, which is more robust to noisy and fragmented text, including TV subtitles~\citep{jorgensen-etal-2016-learning}.
Additionally, we use \texttt{phrasemachine} \cite{Handler2016BagOW} to identify all corpus noun phrases. Finally, we identify player mentions in the transcript text using exact string matches of first, last, and full names to roster information from online archives; these rosters also contain the player's position.\footnote{Roster sources listed in Appendix. We tag first and last name mentions only if they can be disambiguated to a single player in the rosters from opposing teams.} Although we initially had concerns about the reliability of transcriptions of player names, we noticed minimal errors on more common names. Qualitatively, we noticed that even uncommon names were often correctly transcribed and capitalized. We leave a more systematic study for future work.

\subsection{Identifying player race}

Racial identity in the United States is a creation of complex, fluid social and historical processes \cite{omi2014racial}, rather than a reflection of innate differences between fixed groups. Nevertheless, popular \textit{perceptions} of race in the United States and the prior scholarship on racial bias in sports broadcasts which informs our work \cite{rainville1977extent,rada1996color,billings2004depicting,rada2005color} typically assume hard distinctions between racial groups, which measurably affect commentary. In this work, we do not reify these racial categories; we use them as commonly understood within the context of the society in which they arise.

To conduct a large-scale re-examination of this prior work, we must identify whether each player in \ours\ is perceived as white or nonwhite.\footnote{While we use the general term ``nonwhite'' in this paper, the majority of nonwhite football players are black: in 2013, 67.3\% of the NFL was black and most of the remaining players (31\%)
were white~\citep{lapchick20122012}.} Unfortunately, publicly available rosters or player pages do not contain this information, so we resort to crowdsourcing. We present crowd workers on the Figure Eight platform with 2,720 images of professional player headshots from the Associated Press paired with player names. We ask them to ``read the player's name and examine their photo'' to judge whether the player is white or nonwhite. We collect five judgements per player from crowd workers in the US, whose high inter-annotator agreement (all five workers agree on the race for 93\% of players) suggests that their perceptions are very consistent.
Because headshots were only available for a subset of players, the authors labeled the race of an additional 1,948 players by performing a Google Image search for the player's name\footnote{We appended ``NFL'' to every query to improve precision of results.} and manually examining the resulting images. Players whose race could not be determined from the search results were excluded from the dataset.
\section{Analyzing \ours}
\label{sec:analysis}
\begin{figure}[t!]
    \centering
    \includegraphics[width=3in]{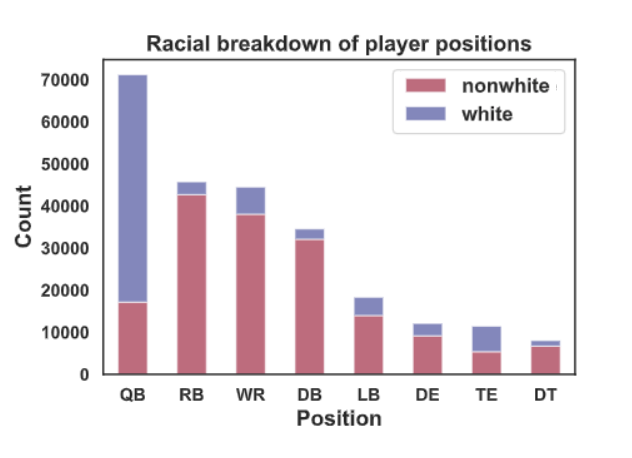}
    \caption{Almost all of the eight most frequently-mentioned positions in \ours\ are heavily skewed in favor of one race.} \label{fig:position}
\vspace{-0.1in}
\end{figure}

\begin{figure}[t!]
    \centering
    \includegraphics[width=3in]{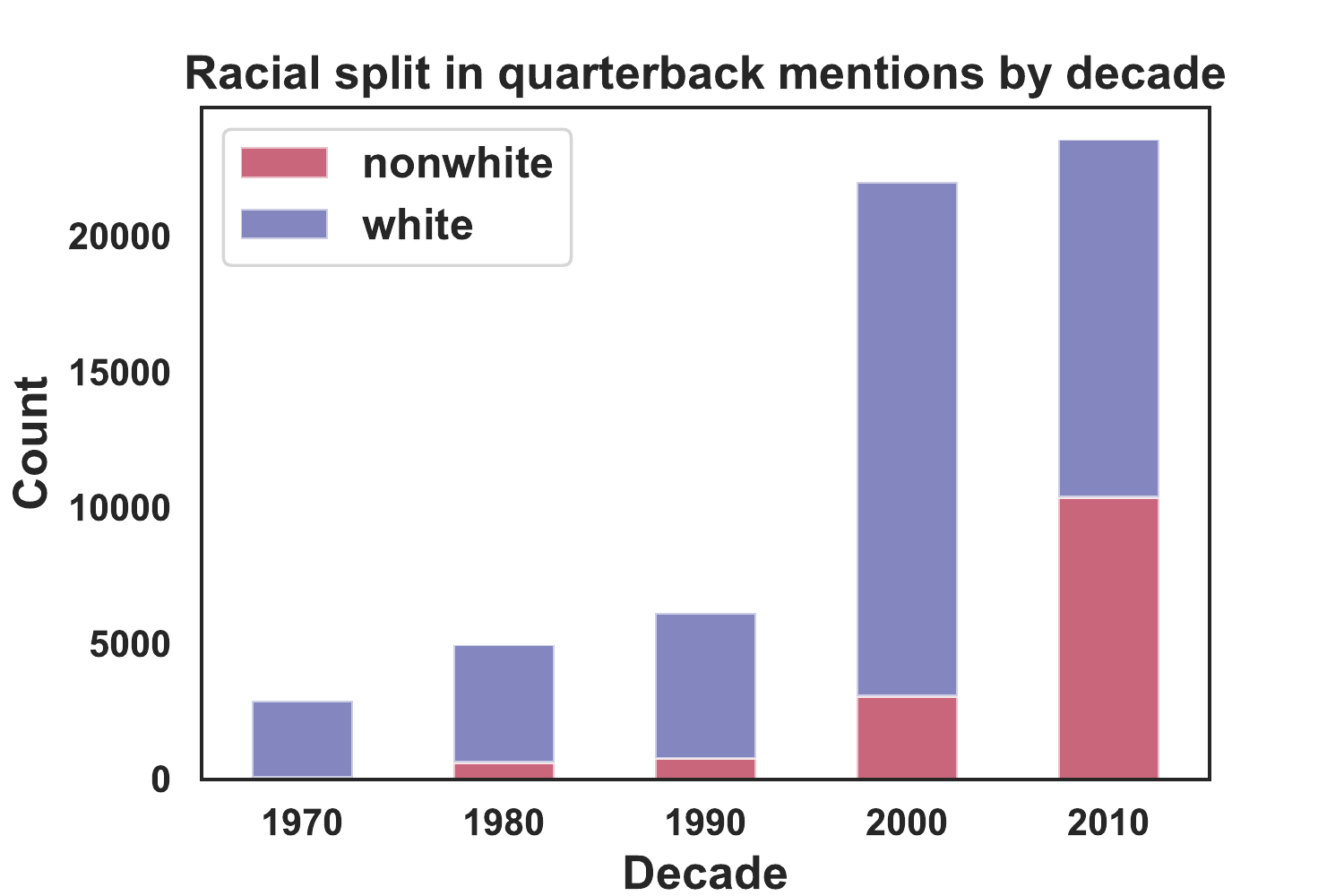}
    \caption{The percentage of nonwhite quarterbacks mentions has drastically increased over time, exemplifying the changing racial landscape in \ours\ across time.} \label{fig:qb_decades}
\vspace{-0.1in}
\end{figure}

We now demonstrate confounds in the data and revisit several established results from racial bias studies in sports broadcasting. 
For all experiments, we seek to analyze the statistics of contextual terms that describe or have an important association with a mentioned player. Thus, we preprocess the transcripts by collecting contextual terms in windows of five tokens around each player mention, following the approach of~\citet{gender:naacl19} for gendered mention analysis.\footnote{If multiple player mentions fall within the same window, we exclude each term to avoid ambiguity.} 

We emphasize that different term extraction strategies are possible, corresponding to different precision--recall tradeoffs. For instance, instead of collecting all terms in a window (high recall) we might instead only collect terms in copular constructions with the entity mention (high precision), such as `devoted' in ``Tebow is devoted''. Because mention detection strategies affect conclusions about bias in \ours, systematically defining, analyzing or even learning different possible strategies offers an exciting avenue for future work.

\subsection{Statistical and linguistic confounds}
\label{subsec:confounds}

Identifying racial bias in football broadcasts presents both statistical and linguistic modeling challenges. Many descriptions of players in broadcasts describe temporary player states (e.g., ``Smith deep in the backfield'') or discrete player actions  (``Ogden with a huge block''), rather than possibly-biased descriptions of players themselves (``Cooper is one scrappy receiver''). Moreover, many players' actions (``passes the ball downfield'') depend on the position they play, which is often skewed by race (Figure~\ref{fig:position}). Furthermore, the racial composition of mentions across different decades can differ dramatically---Figure~\ref{fig:qb_decades} shows these changes for quarterback mentions---which makes the problem even more complex. Modeling biased descriptions of players thus requires disentangling attributes describing shifting, position-dependent player actions on field (e.g., ``Paulsen the tight end with a \textit{fast} catch'') from attributes referring to intrinsic characteristics of individual players (``Paulsen is just so, so \textit{fast}'').

To demonstrate this challenge, we distinguish between per-position effects and racial effects using an additive, log-linear model which represents the log probability that a word or noun phrase $w$ will describe a player entity $e$ as the sum of two learned coefficients, corresponding to two observed covariates. One observed covariate records a player's race and the other a player's position, which allows us to use learned coefficients to represent how much a player's race or position contributes to the chance of observing an $(w,e)$ pair.

Formally, we model such effects using a
sparse MAP estimation variant of SAGE \cite{eisenstein2011sparse}.
We define the binary vector $y_e\in\{0,1\}^J$ 
to represent the observed player covariates of race (white or nonwhite) and position.  For example, component $y_{e,k}$ will be set to 1 if player $e$ is a quarterback and the component $k$ indexes the quarterback covariate; $y_e$ is a concatenation of two one-hot vectors.
We then model 
$ p(w \mid e) \propto \exp\left( \beta_w + (\gamma y_e)_w \right)$, 
with $\beta_w \in \mathbb{R}^{|\mathcal{V}|}$ as  a background distribution over the vocabulary $\mathcal{V}$,
set to empirical corpus-wide word and phrase log-probabilities,
and
$\gamma \in \mathbb{R}^{J \times |\mathcal{V}|}$ as a matrix of feature effects on those  log probabilities.
$\gamma_{j,w}$ denotes the \emph{difference} in log-probability of $w$ for the $j^{th}$ player feature being on versus off.  For example, if $j$ indexes the quarterback covariate and $w$ indexes the word ``tough'', then $\gamma_{j,w}$ represents how much more likely the word ``tough'' is to be applied to quarterbacks over the base distribution. We impose a uniform Laplace prior on all elements of $\gamma$ to induce sparsity, and learn a MAP estimate with the LibLBFGS implementation of OWL-QN, an L1-capable quasi-Newtonian convex optimizer \cite{Andrew2007OWLQN,Okazaki2010LibLBFGS}. We learn from a sample of one million noun phrases and noun tokens from the corpus.

Table \ref{t:covar} shows several highest-valued $\gamma_{j,w}$ for a subset of the $J$ covariates. The adjective ``big'' is predictive of running backs, but refers to an action on the field (``big hole''), not an attribute of running backs. We also find that since ``strong safety'' is a kind of defensive back, the adjective ``strong'' is often associated with defensive backs, who are often nonwhite. In this case, 
``strong'' does not reflect racial bias. Preliminary experiments with per-position mention-level race classifiers,
as per \citet{gender:naacl19}, were also unable
to disentangle race and position.

These results suggest that a more sophisticated approach
may be necessary to isolate race effects from the confounds;
it also raises sharp conceptual questions
about the meaning of race-conditional statistical effects
in social scientific inquiry,
since race is a multifaceted construct
(a ``bundle of sticks,'' as \citet{Sen2016Bundle} argue).
For future work, it may be useful to think of comparisons between otherwise similar players: how do broadcasters differ in their discussions of two players who are both quarterbacks, and who have similar in-game performance, but differ by race?


We now describe two experiments that sidestep some of these confounds,
each motivated by prior work in social science: the first
examines player naming patterns, which are less tied to action on field than player attributes. The other uses words with known sentiment polarity to identify positive and negative attributes, regardless of player position or game mechanics.

\begin{table}[t!]
\small
\begin{tabular}{@{}rl@{}}
\textbf{White} & {\textit{\small long time, official, free safety}} \\ 
\textbf{DB} & {\textit{\small great coverage, strong safety, free safety}}  \\
\textbf{RB} & {\textit{\small big hole, more yards, great block}} \\
\textbf{QB} & {\textit{\small plenty of time, florida state, comfortable}} \\ 
\textbf{WR} & {\textit{\small double coverage, total, wide receivers}} \\ 
\end{tabular} 
\caption{Top terms for the white, defensive back (DB), running back (RB), quarterback (QB),
and wide receiver (WR)
covariates for the log linear model.}
\label{t:covar}
\vspace{-0.1in}
\end{table}

\subsection{Exploring naming patterns} \label{naming}
\emph{Naming patterns} in sports broadcasting---how commentators refer to players by name (e.g., first or last name)---are influenced by player attributes, as shown by prior small-scale studies. For example,~\newcite{koivula1999gender} find that women are more frequently referred to by their first names than men in a variety of sports. \citet{bruce2004marking} discover a similar trend for race in basketball games: white players are more frequently referred to by their last names than nonwhite players, often because commentators believe their first names sound too ``normal''.~\citet{bruce2004marking} further points out that the ``practice of having fun or playing with the names of people from non-dominant racial groups'' contributes to racial ``othering''. A per-position analysis of player mentions in~\ours\ corroborates these findings for all offensive positions (Table~\ref{tab:naming}). 

\begin{table}[t]
\small
\begin{center}
\begin{tabular}{ llrrr } 
\toprule
\bf Position & \bf Race & \bf First & \bf Last & \bf Full  \\ 
\midrule
QB & white & 8.3\% &  20.0\% & 71.7\% \\
QB & nonwhite & 18.1\% &  7.5\% & 74.5\% \\
\midrule
WR & white & 6.9\% &  36.5\% & 56.5\% \\
WR & nonwhite & 11.3\% &  24.1\% & 64.6\% \\
\midrule
RB & white & 10.5\% &  41.2\% & 48.4\% \\
RB & nonwhite & 8.5\% &  35.4\% & 56.1\% \\
\midrule
TE & white & 16.6\% &  18.7\% & 64.7\% \\
TE & nonwhite & 13.8\% &  16.6\% & 69.7\% \\
\bottomrule
\end{tabular}
\end{center}
\caption{White players at the four major offensive positions are referred to by last name more often than nonwhite players at the same positions, a discrepancy that may reflect unconscious racial boundary-marking.}
\label{tab:naming}
\vspace{-0.1in}
\end{table}


\subsection{Sentiment patterns} \label{sentiment}
Prior studies examine the relationship between commentator sentiment and player race:~\citet{rainville1977extent} conclude that white players receive more positive coverage than black players, and~\citet{rada1996color} shows that nonwhite players are praised more for physical attributes and less for cognitive attributes than white ones. 

To examine sentiment patterns within \ours, we assign a binary sentiment label to contextualized terms (i.e., a window of words around a player mention) by searching for words that match those in domain-specific sentiment lexicons from~\newcite{hamilton2016inducing}.\footnote{We use a filtered intersection of lexicons from the NFL, CFB, and sports subreddits, yielding 121 positive and 125 negative words.} 
This method identifies 49,787 windows containing sentiment-laden words, only 12.8\% of which are of negative polarity, similar to the 8.3\% figure reported by~\citet{rada1996color}.\footnote{Preliminary experiments with a state-of-the-art sentiment model trained on the Stanford Sentiment Treebank \cite{peters-etal-2018-deep} produced qualitatively unreliable predictions due to the noise in \ours.} We compute a list of the most positive words for each race ranked by ratio of relative frequencies~\citep{monroe2008fightin}.\footnote{We follow~\citet{monroe2008fightin} in removing infrequent words before ranking; specifically, a word must occur at least ten times for each race to be considered.} A qualitative inspection of these lists (Table~\ref{tab:polarize_terms}) confirms that nonwhite players are much more frequently praised for physical ability than white players, who are praised for personality and intelligence (see Table~\ref{tab:dataset_examples} for more examples). \\\\
\noindent\textbf{Limitations:} 
The small lexicon results in the detection of relatively few sentiment-laden windows; furthermore, some of those are false positives (e.g., ``beast mode'' is the nickname of former NFL running back Marshawn Lynch). The former issue precludes a per-position analysis for all non-QB positions, as we are unable to detect enough sentiment terms to draw meaningful conclusions. The top two rows of Table~\ref{tab:polarize_terms}, which were derived from all mentions regardless of position, are thus tainted by the positional confound discussed in Section~\ref{subsec:confounds}. 
The bottom two rows of Table~\ref{tab:polarize_terms} are derived from the same analysis applied to just quarterback windows; qualitatively, the results appear similar to those in the top two rows. That said, we hope that future work on contextualized term extraction and sentiment detection in noisy domains can shed more light on the relationship between race and commentator sentiment patterns.


\begin{table}[t!]
\small
\begin{center}
\begin{tabular}{ lp{4.6cm} } 
\toprule
\bf Race & \bf Most positive words  \\ 
\midrule
white (all) & \emph{enjoying, favorite, calm, appreciate, loving, miracle, spectacular, perfect, cool, smart}\\
nonwhite (all) & \emph{speed, gift, versatile, gifted, playmaker, natural, monster, wow, beast, athletic}\\
\midrule
white (QBs) & \emph{cool, smart, favorite, safe, spectacular, excellent, class, fantastic, good, interesting}\\
nonwhite (QBs) & \emph{ability, athletic, brilliant, awareness, quiet, highest, speed, wow, excited, wonderful}\\
\bottomrule
\end{tabular}
\end{center}
\caption{Positive comments for nonwhite players (top two rows: all player mentions; bottom two rows: only quarterback mentions) focus on  their athleticism, while white players are praised for personality and intelligence.}
\label{tab:polarize_terms}
\vspace{-0.1in}
\end{table}


\section{Related Work}

\label{sec:related_work}
Our work revisits specific findings from social science (\S\ref{sec:analysis}) on racial bias in sports broadcasts. Such non-computational studies typically examine a small number of games drawn from a single season and rely on manual coding to identify  differences in announcer speech \cite{rainville1977extent,billings2004depicting,rada2005color}. For example, \newcite{rada1996color} perform a fine-grained analysis of five games from the 1992 season, coding for aspects such as players' cognitive or physical attributes.  Our computational approach allows us to revisit this type of work (\S\ref{sec:analysis}) using \ours, without relying on subjective human coding.



Within NLP, researchers have studied gender bias in word embeddings \cite{Bolukbasi2016ManIT,Caliskan183}, racial bias in police stops \cite{Voigt6521} and on  Twitter~\citep{hasanuzzaman2017demographic}, and biases in NLP tools like sentiment analysis systems~\citep{Kiritchenko2018ExaminingGA}. Especially related to our work is that of~\newcite{gender:naacl19}, who analyze mention-level gender bias, and \newcite{tennisgender:IJCAI}, who examine gender bias in tennis broadcasts. 
Other datasets in the sports domain include the event-annotated baseball commentaries of~\newcite{keshet2011ballgame} and the WNBA and NBA basketball commentaries of~\newcite{aull2013fighting}, but we emphasize that \ours\ is the first large-scale sports commentary corpus annotated for race. 
\section{Conclusion}\label{sec:conclusion}


We collect and release \ours\ to support large-scale, longitudinal analysis of racial bias in sports commentary, a major category of mass media. Our analysis confirms the results of prior smaller-scale social science studies on commentator sentiment and naming patterns.
However, we find that baseline NLP methods for quantifying mention-level genderedness~\citep{gender:naacl19} and modeling covariate effects ~\citep{eisenstein2011sparse} cannot overcome the statistical and linguistic confounds in this dataset. We hope that presenting such a technically-challenging resource, along with an analysis showing the limitations of current bias-detection techniques, will contribute to the emerging literature on bias in language. Important future directions include examining the temporal aspect of bias as well as developing more precise mention identification techniques. 

\section*{Acknowledgments}
We thank the anonymous reviewers for their insightful comments,
and Emma Strubell, Patrick Verga, David Fisher, Katie Keith, Su Lin Blodgett, and other members
of the UMass NLP group for help with
earlier drafts of the paper.
This work was partially supported by
NSF IIS-1814955 and a Figure Eight AI for Everyone award.


\bibliographystyle{bib/acl_natbib}
\bibliography{bib/journal-full,bib/sports_bias}


\end{document}


\section*{Appendix}
\label{sec:appendixA}

Here we provide tables that give a much more detailed breakdown of \ours.\ For instance, we show breakdowns by position (Table~\ref{tab:num_position}), time period (Table~\ref{tab:num_games}), and race (Table~\ref{tab:lab_ments_total}). Additionally, we fully specify the transcript and roster collection process.

\begin{table}[h!]
    \begin{center}
    \scalebox{1.0}{%
    \begin{tabular}{ lrrr } 
        \toprule
        Position & white & nonwhite & Total \\ 
        \midrule
            QB & 54.0k & 17.2k & 71.3k \\ 
            RB & 3.1k  & 42.8k & 45.8k \\ 
            WR & 6.5k & 38.1k & 44.6k \\ 
            DB & 2.5k & 32.1k & 34.6k \\ 
            LB & 4.4k & 14.0k & 18.4k \\ 
            DE & 3.0k & 9.2k & 12.2k \\ 
            TE & 6.1k & 5.4k & 11.448 \\ 
            DT & 1.3k & 6.7k & 8.0k \\ 
            OT & 2.7k & 3.3k & 6.0k \\
            K  & 5.7k & 279 & 5.9k \\
            OG & 1.9k & 2.1k & 4.0k \\
            P  & 3.6k & 219 & 3.8k \\
            C  & 1.4k & 254 & 1.6k \\ 
            LS & 92 & 0 & 92 \\ 
            OL & 27 & 38 & 65 \\ 
            DL & 11 & 51 & 62 \\  \bottomrule
    \end{tabular}}
    \end{center}
    \caption{Number of mentions in \ours\  by position. There are 267,778 player mentions in total.}
    \label{tab:num_position}
\end{table}

\begin{table}[h!]
\begin{center}
\scalebox{1.0}{%
\begin{tabular}{ lrrrr } 
\toprule
Years & NFL & NCAA & Total \\ 
\midrule
1960-1969 & 5 & 0 & 5  \\
1970-1979 & 53 & 50 & 103  \\
1980-1989 & 36 & 76 & 112  \\
1990-1999 & 57 & 106 & 163 \\
2000-2009 & 105 & 194 & 299  \\
2010-present & 345 & 428 & 773 \\
\bottomrule
\end{tabular}}
\end{center}
\caption{Number of games in \ours\ by decade}
\label{tab:num_games}
\end{table}

\begin{table}[h!]
 \small
\begin{center}
\scalebox{1.0}{%
\begin{tabular}{ lrrrr } 
\toprule
& \multicolumn{2}{c}{Mentions by Decade}\\
Years & nonwhite & white & Total \\ 
\midrule
1960-1969 & 1.0k & 641 & 1.6k  \\
1970-1979 & 9.4k & 9.2k & 18.6k  \\
1980-1989 & 7.3k & 11.1k & 18.4k  \\
1990-1999 & 9.5k & 18.9k & 28.4k \\
2000-2009 & 18.7k & 35.8k & 54.5k  \\
2010-present & 50.2k & 95.9k & 146.1k \\
\bottomrule
\end{tabular}}
\end{center}
\caption{Number of mentions in \ours\  by decade}
\label{tab:data_stats}
\end{table}

\begin{table}[h!]
 \small
\begin{center}
\scalebox{1.0}{
\begin{tabular}{ lrr  } 
\toprule
League & nonwhite & white  \\ 
\midrule
NFL & 137.4k & 84.0k   \\
NCAA & 34.2k & 12.1k   \\ \midrule
\textbf{Total} & \textbf{171.6k} & \textbf{96.1k}   \\ \bottomrule
\end{tabular}}
\end{center}
\caption{Mentions in \ours\ by race and league}
\label{tab:lab_ments_total}
\end{table}

\begin{table}[h!]
\begin{center}
\scalebox{0.8}{%
\begin{tabular}{ lrrrr } 
\toprule
 & \multicolumn{2}{c}{NFL} & \multicolumn{2}{c}{NCAA} \\
 Years & nonwhite & white & nonwhite & white\\
\midrule 
1960-1969 & 75 & 40 & 0 & 0 \\ 
1970-1979 & 189 & 198 & 18 & 39 \\ 
1980-1989 & 185 & 291 & 15 & 85 \\ 
1990-1999 & 136 & 419 & 27 & 130 \\ 
2000-2009 & 265 & 761 & 76 & 312 \\ 
    2010-present & 481 & 1.3k & 116 & 419 \\ 
\bottomrule
\end{tabular}}
\end{center}
\caption{Number of distinct labeled players}
\label{tab:num_players}
\end{table}

\begin{table}[h!]
\begin{center}
    \scalebox{0.8}{%
    \begin{tabular}{ lrrr } 
        \toprule
        Years & \% white (dataset) & \% white (NFL) \\ 
        \midrule
        1960-1969 & 61.0 & * \\ 
        1970-1979 & 47.1 & * \\ 
        1980-1989 & 38.4 & * \\ 
        1990-1999 & 29.9 & 33.0 \\ 
        2000-2009 & 40.5 & 30.6 \\ 
        2010-present & 28.1 & 29.9 \\ 
        \bottomrule
        \end{tabular}}
    \end{center}
    \caption{Percentage of mentions by race and decade in \ours\ compared with the actual race distribution of players in the NFL by decade \cite{lapchick2017}. The distributions are very similar. *\citet{lapchick2017} does not provide data for this time span.}
    \label{tab:data_stats}
\end{table}

\subsection*{Transcript collection: additional details} 
We collect YouTube videos from the following YouTube channels, which aggregate football games \emph{Mark Jones},
\emph{NFL Full Games 2018 / 2019},
\emph{Adrián GTZ Montoya},
\emph{NFL Full Games},
\emph{Bart Simpson},
\emph{Bryan Mears},
\emph{NFL},
\emph{Alex Roberts},
\emph{Sports},
\emph{Danger zone},
\emph{Nittany 96}. (Some channels publish dedicated lists of football games, amid other content. Other channels post football games only).

After downloading videos, we identified the teams playing and the year of each video by matching strings in video titles. In isolated cases when string matching failed, we manually identified the teams and year from the video itself.

We remove stop words and entities from our original text before processing mentions using the NLTK English stopwords list.

\subsection*{Roster collection: additional details} 

We collected rosters for all NFL teams from 1960 to the present from \url{footballdb.com}. Because some players have the same name, we used the player name-position pair (e.g. Tom Brady, QB) to identify a unique entity in our dataset. 

We collect NCAA rosters from the college football page \url{ sports-reference.com}, downloading rosters for the 290 available schools in the NCAA from 1869 to the present.

We lower case all names on rosters. We also remove periods, apostrophes and hyphens within names (e.g., Odell Beckham, Jr., Ra'shede Hageman, Dominique Rodgers-Cromartie).

In total, including the mentions for which we could not acquire racial metadata, we gathered a total of 545,232 labeled and non labeled player mentions, 265,879 NCAA mentions and 279,353 NFL mentions. The higher number of NFL mentions (despite fewer NFL games) is due to incomplete roster information for NCAA teams (some years are incomplete, some years missing altogether). For the players we were able to acquire race labels for, 137,428 nonwhite and 84,038 white mentions are collected from NFL games and 34,196 nonwhite and 12,116 white mentions are collected NCAA games.

\bibliographystyle{bib/acl_natbib}
\bibliography{bib/journal-full,bib/sports_bias}